\documentclass{article}

\usepackage[preprint]{neurips_2023}

\usepackage[utf8]{inputenc} 
\usepackage[T1]{fontenc}    
\usepackage{hyperref}       
\usepackage{url}            
\usepackage{booktabs}       
\usepackage{amsfonts}       
\usepackage{nicefrac}       
\usepackage{microtype}      
\usepackage{xcolor}         

\usepackage{amsthm}
\usepackage[utf8]{inputenc} 
\usepackage[T1]{fontenc}    
\usepackage{hyperref}       
\usepackage{url}            
\usepackage{booktabs}       
\usepackage{amsfonts}       
\usepackage{nicefrac}       
\usepackage{microtype}      
\usepackage{xcolor}         
\usepackage{color, colortbl}
\usepackage{multirow}
\usepackage{amsmath}
\definecolor{dark-blue}{rgb}{0.15,0.15,0.4}
\definecolor{Gray}{gray}{0.9}

\usepackage{sistyle}
\SIthousandsep{,}
\usepackage{halloweenmath}
\usepackage{algorithm,algorithmic}
\usepackage{graphicx}
\usepackage{wrapfig}

\usepackage{caption}
\usepackage{subcaption}
\usepackage[nameinlink,capitalise,noabbrev]{cleveref}

\hypersetup{
    colorlinks=true,
    linkcolor=blue,
    filecolor=magenta,      
    urlcolor=cyan,
    citecolor=blue,
    pdftitle={Battle of the Backbones},
    pdfpagemode=FullScreen,
}

\title{Generating Potent Poisons and Backdoors from Scratch with Guided Diffusion}

\author{
Hossein Souri$^1$\thanks{Correspondence to \url{hsouri1@jhu.edu}. Johns Hopkins University$^1$, University of Maryland$^2$, Google$^3$, ELLIS Institute T\"ubingen \& MPI Intelligent Systems, T\"ubingen AI Center$^4$, and New York University$^5$.}\\
\And
Arpit Bansal$^2$
\And
Hamid Kazemi$^2$
\And
Liam Fowl$^3$
\And
Aniruddha Saha$^2$
\And
Jonas Geiping$^4$
\And
Andrew Gordon Wilson$^5$
\And
Rama Chellappa$^1$
\And
Tom Goldstein$^{2}$
\And
Micah Goldblum$^5$
}

\begin{document}

\maketitle

\begin{abstract}
Modern neural networks are often trained on massive datasets that are web scraped with minimal human inspection.  As a result of this insecure curation pipeline, an adversary can poison or backdoor the resulting model by uploading malicious data to the internet and waiting for a victim to scrape and train on it.  Existing approaches for creating poisons and backdoors start with randomly sampled clean data, called base samples, and then modify those samples to craft poisons.  However, some base samples may be significantly more amenable to poisoning than others. As a result, we may be able to craft more potent poisons by carefully choosing the base samples.  In this work, we use guided diffusion to synthesize base samples from scratch that lead to significantly more potent poisons and backdoors than previous state-of-the-art attacks. Our Guided Diffusion Poisoning (GDP) base samples can be combined with any downstream poisoning or backdoor attack to boost its effectiveness. Our implementation code \mbox{is publicly available at: \url{https://github.com/hsouri/GDP}}.
\end{abstract}    
\section{Introduction}
\label{sec:intro}

Large-scale neural networks have seen rapid improvement in many domains over recent years, enabled by web-scale training sets.  These massive datasets are collected using automated curation pipelines with little to no human oversight.  Such automated pipelines are vulnerable to data tampering attacks in which malicious actors upload harmful samples to the internet that implant security vulnerabilities in models trained on them, in the hopes that a victim scrapes the harmful samples and incorporates them in their training set. For example, \emph{targeted data poisoning attacks} cause the victim model to misclassify specific test samples \citep{shafahi2018poison, geiping2020witches}, while \emph{backdoor attacks} manipulate victim models so that they only misclassify test samples when the samples contain a specific backdoor trigger \citep{gu2017badnets, turner2018clean, saha2020hidden, souri2022sleeper}.

Typical data poisoning and backdoor attacks begin with randomly selected \emph{base samples} from clean data, which the attacker perturbs to minimize a poisoning objective.  The attacker often constrains the perturbations to be small so that the resulting poisons look similar to the original base samples and still appear correctly labeled \citep{geiping2020witches, souri2022sleeper}.  However, a real world attacker may not be constrained around a particular set of randomly chosen base samples.  Indeed, \citet{souri2022sleeper} show that some base samples are far more effective for crafting backdoor poisons than others.  Namely, \citet{souri2022sleeper} select base samples which achieve a high gradient norm, $\|\nabla_{\theta}\ell(\theta)\|_{2}$, where $\ell(\theta)$ denotes the training loss of a classifier with parameters $\theta$.  Such a selection strategy enables more potent poisons, but the base samples are still limited to clean samples chosen from a limited dataset.

\begin{figure*}[t]
    \centering
    \includegraphics[width=\textwidth]{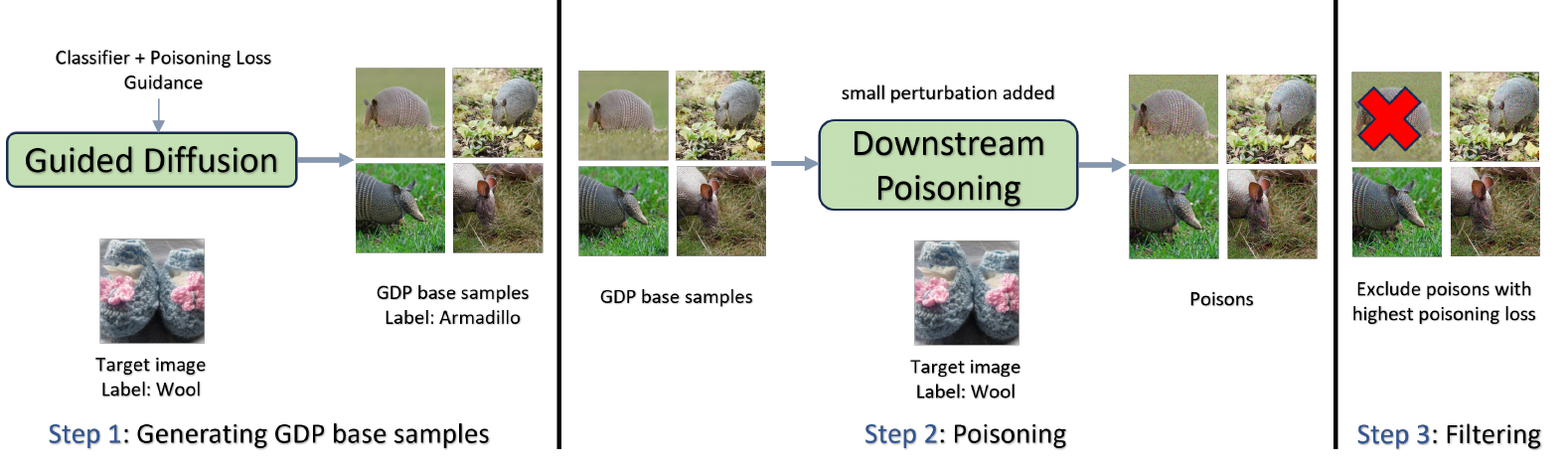}
    \caption{\textbf{Schematic of Guided Diffusion Poisoning (GDP).} GDP contains three stages: \textbf{(1)} generate base samples with a diffusion model weakly guided using a poisoning loss; \textbf{(2)} use the base samples as initialization for a downstream poisoning algorithm; \textbf{(3)} select poisons with the lowest poisoning loss and include them in the poisoned training set.}
    \label{fig:teaser}
\end{figure*}

\begin{figure}[t!]
    \centering
    \includegraphics[width=\linewidth]{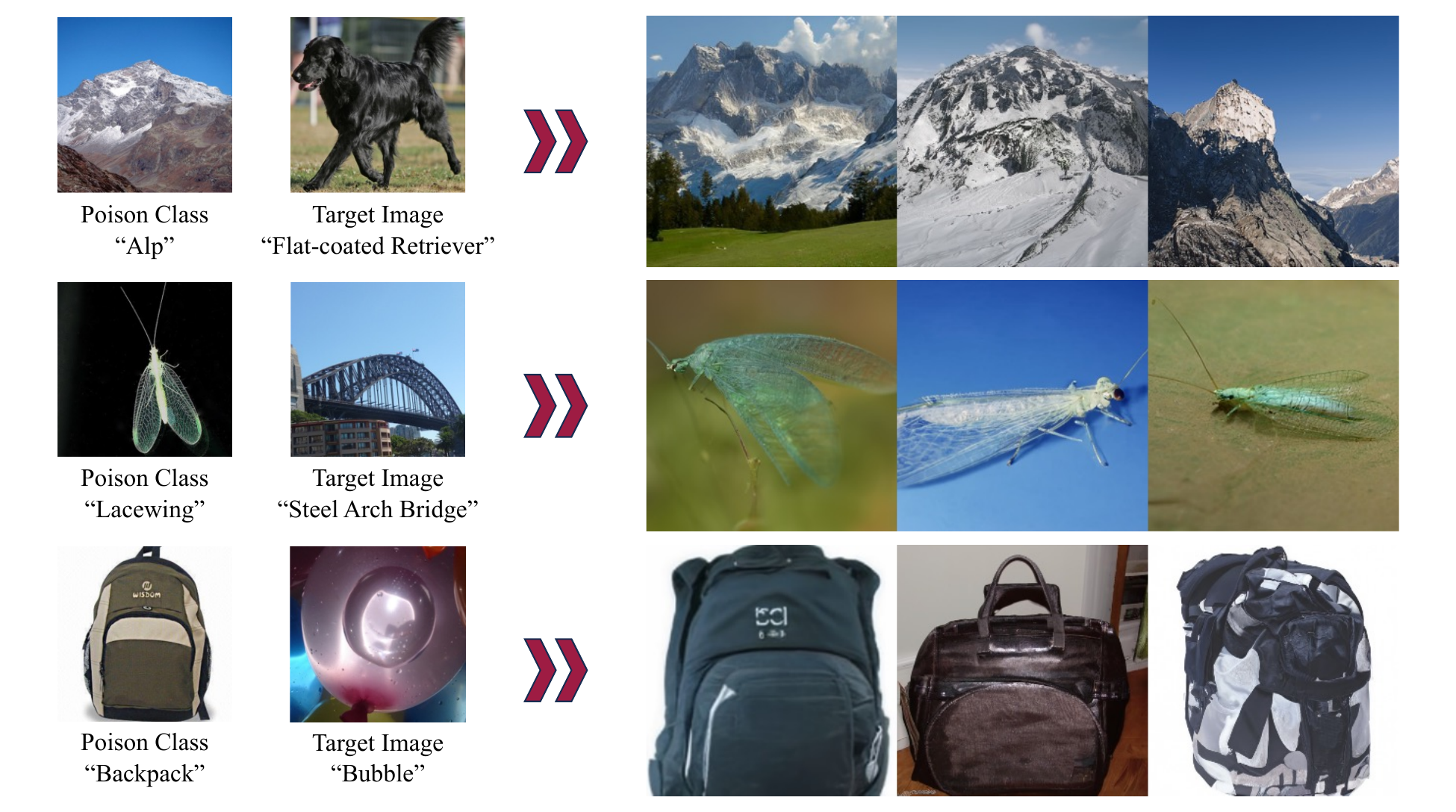}
    \caption{\textbf{GDP base samples are clean-label and high quality (ImageNet).}  In each panel, the leftmost column contains a random sample from the poison class, the second column contains the target image, and the subsequent three columns contain GDP base samples. Experiments conducted using the Witches' Brew gradient-matching objective with a ResNet-18 model on ImageNet over randomly sampled poison class and target image pairs. Additional visualizations are found in \Cref{app:vis}.}
    \label{fig:imagenet_vis}
\end{figure}

\begin{figure}[pt!]
    \centering
    \includegraphics[width=\linewidth]{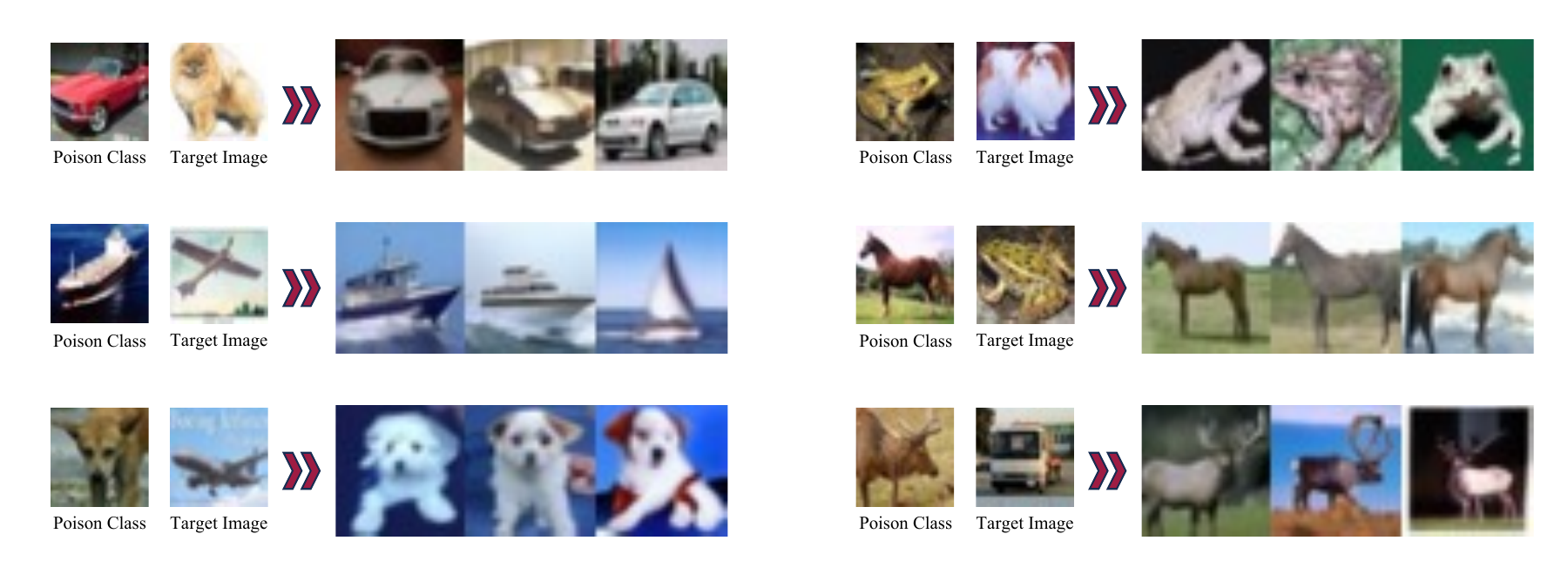}
    \caption{\textbf{GDP base samples are clean-label and high quality (CIFAR-10).}  In each panel, the leftmost column contains a random sample from the poison class, the second column contains the target image, and the subsequent three columns contain GDP base samples. Experiments conducted using the Witches' Brew gradient-matching objective with a ResNet-18 model on CIFAR-10 over randomly sampled poison class and target image pairs. Additional visualizations are found in \Cref{app:vis}.}
    \label{fig:cifar_vis}
\end{figure}

\begin{figure*}[t!]
    \centering
    \includegraphics[width=\linewidth]{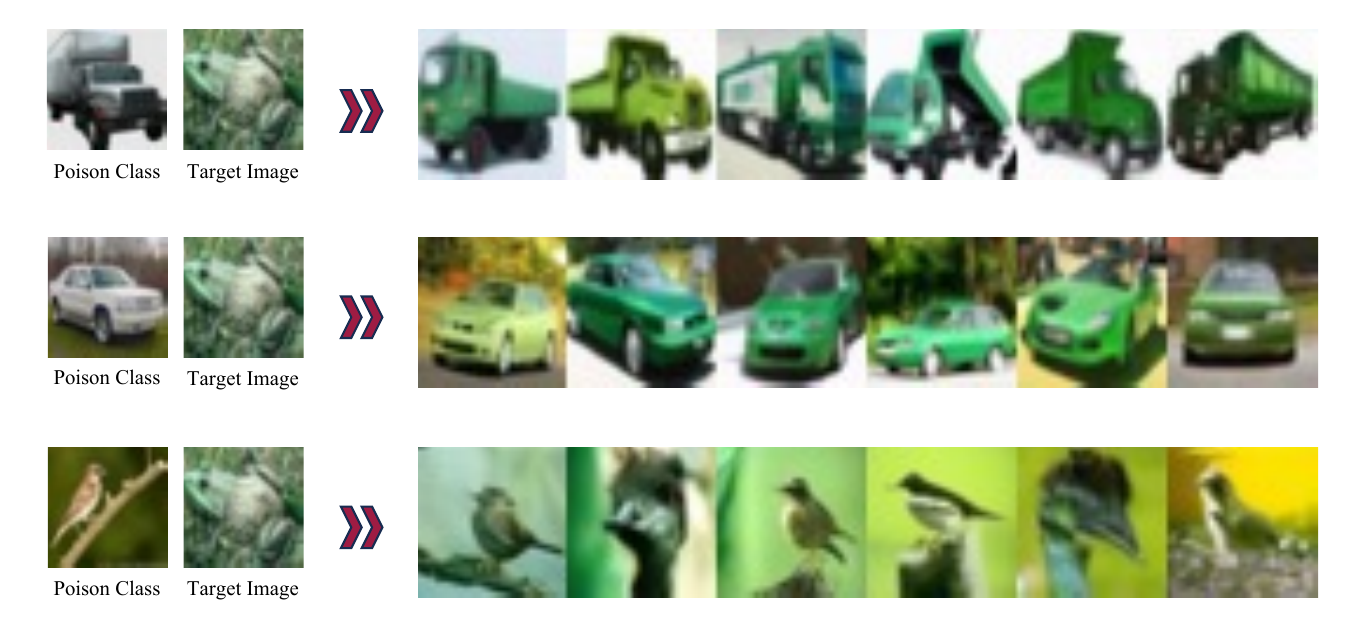}
    \caption{\textbf{GDP produces base samples that look like the target image while still remaining in the poison class.} We generate base samples from different poison classes while the target image is fixed.  We see that all resulting GDP base samples contain similar colors to the target image but remain clean-label.  Experiments conducted on the CIFAR-10 dataset using the Witches' Brew poisoning objective along with a ResNet-18 model.}
    \label{fig:cifar_frog}
\end{figure*}

In this work, we use diffusion models to synthesize base samples from scratch that enable especially potent attacks.  Rather than filtering out existing natural data, synthesizing the samples from scratch allows us to optimize them specifically for the poisoning objective.  By weakly guiding the generative diffusion process using a poisoning objective, we craft images that are simultaneously near potent poisons while also looking precisely like natural images of the base class.  Then, we can use existing poisoning and backdoor attack algorithms on top of our diffusion-generated base poisons to amplify the effectiveness of the downstream attacks, surpassing previous state-of-the-art for both targeted data poisoning and backdoor attacks. See \Cref{fig:teaser} for a schematic of our method and Figures \ref{fig:imagenet_vis}, \ref{fig:cifar_vis}, \ref{fig:cifar_frog} for corresponding example base images.

We use our Guided Diffusion Poisoning (GDP) base samples in combination with both targeted data poisoning and backdoor attacks \citep{geiping2020witches, souri2022sleeper}.  In all cases, we boost the performance of existing state-of-the-art attacks by a wide margin while preserving image quality and crucially without producing mislabeled poisons, measured through a human study.  These resulting attacks bypass all $7$ defenses we test, and they are successful in the hard black-box setting.  Our approach also enables successful attacks with smaller perturbation budgets on the downstream poisoning algorithm and inserting far fewer poison images into the victim's training set than previous state-of-the-art attacks.

Our contributions are summarized as follows:
\begin{itemize}
\item We devise a method, GDP, for synthesizing clean-label poisoned training data from scratch with guided diffusion.
\item Our approach achieves far higher success rates than previous state-of-the-art targeted poisoning and backdoor attacks, including in scenarios where the attacker is only allowed to poison a very small proportion of training samples.
\item We show how GDP can be used to significantly boost the performance of previous data poisoning and backdoor attack algorithms.
\end{itemize}

\section{Related Work}
\label{sec:related_work}

\subsection{Data Poisoning and Backdoor Attacks}

Data poisoning attacks can be roughly grouped by their threat model and the adversary's objective. Early poisoning attacks demonstrated the vulnerability of simple models, like linear classifiers and logistic regression, to malicious training data modifications \citep{biggio2012poisoning, munoz2017towards, steinhardt2017certified, goldblum2022dataset}. These early poisoning attacks often focused on degrading the overall accuracy of the victim model, and occasionally employed label flips in addition to data modifications. Such an attack is \emph{dirty} label, as it assumes the attacker has some control over the victim's labeling scheme. In contrast, \emph{clean} label attacks do not assume control over the victim's labeling method and usually modify data in a visually minimal way - maintaining their original semantic label. This objective (degrading overall victim accuracy) is sometimes referred to as \emph{indiscriminate} poisoning or an \emph{availability} attack. 

Newer availability poisoning attacks have demonstrated the ability to degrade the accuracy of modern deep networks, but often require modifying a large proportion of the training data \citep{huang2021unlearnable, fowl2021adversarial}. These attacks often rely on creating ``shortcuts'' for the victim model to minimize training loss without learning meaningful features of the clean data distribution. 

In contrast to availability attacks, \emph{targeted} poisoning attacks aim to cause a victim model to misclassify a particular target data point, or a small set of target data points, at inference time. Several works successfully poisoned deep neural networks in a transfer learning setting wherein a backbone is frozen, and a linear layer is retrained on top of the feature extractor \citep{shafahi2018poison, zhu2019transferable, aghakhani2021bullseye}. Newer, more powerful attacks successfully poison victim models trained from \emph{scratch} by using $\ell_\infty$ bounded perturbations \citep{geiping2020witches, huang2020metapoison}. However, these attacks often require poisoning a non-negligible amount of training data - for example, \citet{geiping2020witches} requires several hundred poisons to reliably cause misclassification of a single target CIFAR-10 \citep{krizhevsky2009learning} image at a relatively large $\varepsilon=16/255$ perturbation bound \citep{geiping2020witches}.

Instead of fixing particular samples to target with poisoning attacks, another class of attacks, known as \emph{backdoor} attacks, aim to poison a victim so that \emph{any} sample, with the addition of a trigger, will be misclassified at inference time. Triggers can be additive colorful patches, or small pixel modifications. Early attacks showed backdoor vulnerability in several settings, including transfer learning, and, like their availability attack counterparts, often employed label flips as an attack tool \citep{gu2017badnets, chen2017targeted}. More recent backdoor attacks have advanced the capability of backdoor attacks that do not rely on any label flips. Such attacks have proven successful in transfer learning settings \citep{saha2020hidden}, as well as from-scratch settings \citep{souri2022sleeper}. Recent clean-label attacks can operate with a hidden trigger - a trigger which is not present in any of the poisoned training data, but still effectively induces targeted misclassification when applied at inference time. However, it is worth noting that existing backdoor attack success deteriorates as the number of poisons reaches below $100$. For example, \citet{souri2022sleeper} achieve just over $10\%$ attack success rate when $25$ poisons are included in the victim's training data. 

This class of clean-label, hidden trigger backdoor attacks, along with similar targeted poisoning attacks, can be quite pernicious as they generally require a smaller proportion of data to be modified by the attacker (compared to availability attacks), and are harder to detect by hand inspection of the model's performance on holdout sets. Thus, in this work, we focus on pushing the limits of targeted and backdoor attacks in the \emph{low poison budget} regime wherein a victim might only scrape a handful of samples poisoned by a malicious attacker.

\subsection{Guidance in Diffusion Models}

Diffusion models \citep{song2019generative, DDPM_Ho2020} in machine learning have seen a remarkable evolution, particularly in their application to image generation tasks. These models, which simulate the process of transforming a random noise distribution into a specific data distribution, have incorporated various guidance mechanisms to direct this transformation process more precisely. The concept of condition \citep{ho2022classifierfree,bansal2022colddiff,nichol2021glide,whang2022debluriter,wang2022segmentation_diffusion,li2023gligen,zhang2023controlnet} or guidance \citep{dhariwal2021diffusion,bahjat2022ddrm,wang2022zeroshot_diffusion,chung2022general_inverse,lugmayr2022inpaint_diffusion,chung2022maniold_diffusion,graikos2022diffusion_plug_and_play, bansal2023universal} in diffusion models is crucial for achieving outputs that adhere to specific characteristics or criteria, a necessity in applications demanding high precision.

Initially, guidance within diffusion models was predominantly achieved through two methods: classifier guidance \citep{dhariwal2021diffusion} and classifier-free guidance \citep{ho2022classifierfree}. Classifier guidance \citep{dhariwal2021diffusion} involves training a separate classifier, adept at handling noisy image inputs. This classifier generates a guidance signal during the diffusion process, steering the generative model toward desired outcomes. However, this method necessitates the training of a specialized classifier, often a resource-intensive task. In contrast, classifier-free guidance \citep{ho2022classifierfree} internalizes the guidance mechanism within the model's architecture. This method, while eliminating the need for an external classifier, comes with its limitation: once trained, its adaptability is restricted, unable to accommodate different classifiers or evolving guidance criteria.

To address these constraints, the Control Net \citep{zhang2023controlnet} approach was introduced, representing a significant development in guided diffusion models. Control Net, though requiring less training than traditional classifier guidance, still necessitates some degree of model training. Moreover, its utility is predominantly confined to image-to-image signal guidance, limiting its scope. On the other hand, Universal guidance \citep{bansal2023universal} takes a different approach and completely eschews the need for training new models or classifiers for guidance. Instead, it utilizes signals that can be derived from clean images, employing existing models or loss functions. This strategy significantly enhances the flexibility and efficiency of guidance in diffusion models. However, compared to Control Net, it is computationally expensive during inference.

\section{Background}
\label{sec:background}

\subsection{Poisoning Setup}
Poisoning deep networks trained from scratch is a problem that cannot be solved exactly. This is due to the bi-level nature of the optimization problem, where the attacker tries to find perturbations $\delta = \{\delta_i\}_{i=1}^N$ constrained by conditions $\mathcal{C}$ (usually an $\ell_\infty$ bound) to minimize an adversarial objective $\mathcal{A}$. Formally stated, the attacker solves the following optimization problem:\\
\begin{equation*}
    \min_{\delta \in \mathcal{C}} \mathcal{A}(f_{\theta_*}) \quad \text{s.t.}
\end{equation*}
\begin{equation*}
  \theta_* \in \text{argmin}_\theta \bigg[ \frac{1}{|\mathcal{T}|} \sum_{i=1}^{|\mathcal{T}|} \mathcal{L}(f_\theta (x_i + \delta_i), y_i)  \bigg],
\end{equation*}

where $f_{\theta_*}$ denotes a victim model which itself is trained on the poisons included in the victim's training set, $\mathcal{T}$. Note that $\delta_i = \Vec{0}$ for any $i > N$ (included for simplicity of presentation).

Because of the complexity of this problem, we need to use approximations to solve it. The current gold standard for clean-label poisoning for both targeted and backdoor attacks generally involves \emph{gradient alignment} \citep{geiping2020witches, souri2022sleeper}. Gradient alignment was introduced as a method to poison victim models in \citet{geiping2020witches}, where, in plain words, the attacker imperceptibly modifies poisons so that the gradient of a surrogate model evaluated at these poisons \emph{aligns} with an \emph{adversarial gradient}. In \citet{geiping2020witches}, this adversarial gradient was generated to cause targeted misclassification of a particular image. However, the strategy also generalizes to other poisoning objectives, like a hidden trigger backdoor attack \citep{souri2022sleeper}. More formally, let $f_\theta$ be a surrogate model available to the attacker, $P = \{x_i, y_i\}_{i=1}^N$ be a set of data samples with ground-truth labels which the attacker poisons, and $\mathcal{L}$ be the standard categorical cross-entropy loss, then poisons are crafted by minimizing the alignment objective $\mathcal{O}$ (over the perturbations $\{\delta_i\}$):\\
\begin{equation}
    \mathcal{O} = 1 - \frac{1}{N}\sum_{i=1}^N{\frac{\langle \nabla_\theta \mathcal{L}(f_\theta(x_i + \delta_i), y_i), \nabla_\theta \mathcal{A} \rangle}{||\nabla_\theta \mathcal{L}(f_\theta(x_i + \delta_i), y_i)||\cdot ||\nabla_\theta \mathcal{A}||}}.
\end{equation}

In the case of targeted poisoning, the attacker wishes to induce misclassification of a target image $x'$ into an incorrect class $y'$, so they set
\begin{equation*}
    \mathcal{A} = \mathcal{L}(f_\theta (x', y')).
\end{equation*}

For backdoor attacks, the attacker aims for \emph{any} image which has trigger $t$ applied to be misclassified with label $y'$, so, they choose an adversarial objective\\
\begin{equation*}
    \mathcal{A} = \mathbb{E}_{x \sim D} \mathcal{L}(f_\theta (x + t), y'),
\end{equation*}

where the expectation is approximated over a handful of samples available to the adversary. In practice, the label $y'$ is usually called the \emph{target class}, and the perturbations are usually applied to the \emph{source class}. As an example, an attacker may modify images of dogs imperceptibly so that, when trained on, these dogs cause a victim network to correctly classify a clean dog at inference time, but misclassify that same dog when a trigger patch is applied. 

\subsection{Universal Guidance}
Diffusion models have become pivotal in generative modeling, especially for image generation. They operate through a \(T\)-step process involving both forward and reverse phases. The forward phase incrementally infuses Gaussian noise into an original data point \(x_0\), as described by the equation:\\
\begin{equation}
    x_t = \sqrt{\alpha_t} x_0 + \sqrt{1 - \alpha_t} \epsilon, \quad \epsilon \sim \mathcal{N}(0, \mathbf{I}),
\end{equation}

where \(\epsilon\) represents a standard normal random variable, and \(\alpha_t\) are predefined noise scales.

Conversely, the reverse phase methodically removes noise, aiming to retrieve \(x_0\). This is achieved via a denoising network \(\epsilon_{\theta}\), trained to estimate the noise \(\epsilon\) in \(x_t\) at any step \(t\):\\
\begin{equation}
    \epsilon_{\theta}(x_t, t) \approx \epsilon = \frac{x_t - \sqrt{\alpha_t} x_0}{\sqrt{1 - \alpha_t}}.
\end{equation}

The Denoising Diffusion Implicit Model (DDIM) \citep{DDIM_Song2021} is a prominent reverse process method. It starts by estimating a clean data point \(\hat{x}_0\):\\
\begin{equation}
    \hat{x}_0 = \frac{x_t - \sqrt{1 - \alpha_t} \epsilon_{\theta}(x_t, t)}{\sqrt{\alpha_t}}.
\end{equation}

Then, \(x_{t-1}\) is sampled from \(q(x_{t-1} | x_t, \hat{x}_0)\), replacing \(x_0\) with \(\hat{x}_0\) in the sampling formula:\\
\begin{equation}
    \hat{x}_{t-1} = \sqrt{\alpha_t} \hat{x}_0 + \sqrt{1 - \alpha_t} \epsilon_{\theta}(x_t, t).
\end{equation}

The universal guidance algorithm \citep{bansal2023universal} leverages \(\hat{x}_0\) to create a guidance signal. It includes forward and backward guidance; the latter modifies \(\hat{x}_0\), while the former adapts classifier guidance to suit any general guidance function by calculating the gradient relative to \(x_t\). 

In forward guidance, an external function \(f\) and a loss function \(\ell\) guide image generation. This process is encapsulated by:
\begin{equation}
\label{eq:forward_guidance}
    \hat{\epsilon}_{\theta} (x_t, t) = \epsilon_{\theta} (x_t, t) + s(t) \cdot \nabla_{x_t} \ell (c, f(\hat{x}_0)),
\end{equation}

where \(s(t)\) adjusts the guidance strength at each step. This approach ensures images align with the guidance while maintaining a trajectory within the data manifold. Additionally, the paper introduces universal stepwise refinement, a technique that repeats steps to align gradients, enhancing guidance and image fidelity.

\section{Method}
\label{sec:method}

\subsection{Threat Model}
\label{sec:threat_model}

We adhere to the standard threat model commonly employed in targeted data poisoning and backdoor attacks \citep{geiping2020witches,souri2022sleeper,saha2020hidden}. Assuming the attacker possesses access to the training set, it can manipulate a small subset of the training data by perturbing images within an $\ell_{\infty}$-norm bound from clean non-perturbed images. In our baseline experiments, we consider a gray-box scenario, where the attacker is aware of the model architecture of the victim but lacks knowledge of the parameters. In addition, we assume the attack must be clean-label. Our backdoor attacks operate on a one-to-one basis, focusing on manipulating a single class and being triggered solely by a single patch during testing. The trigger, which is a random patch in our setup, must remain \textit{hidden} from the training set and is chosen arbitrarily. After crafting the poisons, the victim trains their model from scratch on the poisoned training data. We then measure the \textit{Poison Success} rate and \textit{Attack Success Rate} for targeted data poisoning and backdoor attacks, respectively. Further elaboration on the threat model is available in \Cref{app:exp_setup}.

\subsection{Attack Workflow}
We weakly guide diffusion models with a gradient-matching poisoning objective to create base samples that lie very close to potent poisons, but simultaneously preserve image quality and appear similar to samples drawn from the training distribution.  The latter properties ensure that the resulting poisons will be difficult to detect via visual inspection. Diffusion-generated base samples must also be \emph{clean-label} meaning that their assigned labels are aligned with the semantic content of the images.  We then use these base samples as an initialization for existing data poisoning and backdoor attack algorithms to boost their effectiveness. We now detail our simple three-step process for generating base samples and using them for downstream poison and backdoor generation:


\textbf{(1) Generating base samples with guided diffusion:} We use a diffusion model to generate base samples, leveraging a pretrained classifier.  On each step of diffusion, we compute a cross-entropy loss using the probability the classifier assigns to the noisy image iterate being in the poison class.  We include this classification loss in the universal guidance algorithm.  For ImageNet \citep{deng2009imagenet} experiments, we use the classifier-guided diffusion method proposed in \citet{dhariwal2021diffusion} to generate base samples.  Guidance increases the extent to which the generated image looks like the poison class according to the classifier.  We can adjust the effect of class-conditioning on the generated samples by scaling the guidance strength.  In practice, our procedure will reliably produce clean-label base samples (see \Cref{sec:human_eval}).

In addition to the classifier loss, we also add a data poisoning or backdoor loss function in the universal guidance algorithm when generating base samples, but with a very low guidance strength to preserve image quality and prevent the generated images from moving into the target class. In our experiments, we specifically use the Witches' Brew \citep{geiping2020witches} and Sleeper Agent \citep{souri2022sleeper} gradient matching loss functions for targeted data poisoning and backdoor attacks, respectively. We modify the gradient matching objectives from both of these works to optimize per-sample gradient matching instead of matching the average gradient computed over a batch of images.  We find that this modification improves image quality and yields effective base samples in practice.

\textbf{(2) Initializing poisoning and backdoor attacks with GDP base samples:}
After generating base samples, we use them as an initialization for state-of-the-art targeted data poisoning and backdoor attacks. Since these attacks typically constrain their perturbations via $\ell_\infty$ norm in order to ensure that the images are high quality and preserve the poison label, we constrain the attacks accordingly around the initial base samples.  The validity of this procedure depends on the quality and clean-label of the generated base samples, which we will verify. 
 We observe that this procedure leads to poisons which are much more potent and result in stronger attacks than ones which begin with randomly selected clean images.

\textbf{(3) Filtering poisons:}  Given the randomness inherent to diffusion modeling, some base samples do not result in effective poisons. After generating a collection of poisoned training samples, we filter them by selecting the samples that exhibit the lowest value of the corresponding downstream data poisoning or backdoor loss.  We select a number of samples equal to the poisoning budget and replace them with randomly selected clean samples from the victim's training set, thereby maintaining the original size of the training set.  We exclude and ignore the poisoned samples with high poisoning loss.

\Cref{alg1} provides a summary of the proposed GDP attack. It outlines the key steps of the attack, offering an overview of how it is carried out. For visualization of the GDP base samples and final perturbed poisons on CIFAR-10 and ImageNet for poisoning and backdoor attacks, please refer to \Cref{app:vis}.

\begin{algorithm}[t]
\small
    \caption{Guided Diffusion Poisoning (GDP)}
    \label{alg1}
    \begin{algorithmic}[1]
        \renewcommand{\algorithmicrequire}{\textbf{Input:}}
        \renewcommand{\algorithmicensure}{\textbf{Begin:}}
        \REQUIRE Diffusion model $G$, surrogate model $F_\theta$, downstream poisoning loss $L$, poison class $c$, number of base samples $N$, poison budget $P \leq N$
        \ENSURE
        \FOR {$i$ = 1,\ 2,\ ...\ ,\ $N$}
            \STATE Generate base sample $b_i$ with class label c from $G$ using \\
            forward universal guidance as in \Cref{eq:forward_guidance}
        \ENDFOR
        \STATE Randomly initialize perturbations $\delta_{i=1}^N$
        \STATE Calculate $\delta_{i=1}^N$ by optimizing $L(F_\theta, (b_i + \delta_i, c)_{i=1}^N)$
        \STATE Select $P$ poisons from $(b_i + \delta_i, c)_{i=1}^N$ with lowest value of $L(F_\theta, (b_i + \delta_i, c))$
        \STATE Replace $P$ randomly selected training images from class $c$ with $(b_i + \delta_i, c)_{i=1}^P$
    \end{algorithmic} 
\end{algorithm}

\section{Experimental Evaluations}
\label{sec:experiments}

\begin{table*}[t]
\caption{\textbf{Targeted data poisoning. GDP achieves a far higher success rate than existing targeted data poisoning attacks, even with only a small budget}. Experiments are conducted on CIFAR-10 with ResNet-18 models. Perturbations are bounded by $\ell_{\infty}$-norm $16/255$. Poison budget refers to the number of images poisoned. We see that injection of only 25-50 poisoned samples is enough for the attack to be effective.} 
\label{tab:baseline_cifar_poisoning}

\centering 
\resizebox{\textwidth}{!}{
    \begin{tabular}{l@{\hspace{7mm}}c@{\hspace{7mm}}c@{\hspace{7mm}}c@{\hspace{7mm}}c} 
    \toprule 
    \textbf{Poison budget} &\textbf{Poison Frogs} & \textbf{Bullseye} &\textbf{Witches' Brew} & \textbf{GDP (ours)} \\
    \midrule
    $25$ images ($0.05\%$) & $0\%$ & $0\%$ & $0\%$ & $\textbf{30\%}$  \\
    $50$ images ($0.1\%$) & $0\%$ & $0\%$ & $20\%$ & $\textbf{70\%}$  \\
    \bottomrule 
    \end{tabular}
}

\end{table*}

In this section, we evaluate the proposed poisoning pipeline for poisoning image classification models trained on CIFAR-10 and ImageNet. We perform experiments on both targeted data poisoning, following the threat model of \citet{shafahi2018poison}, and backdoor attacks, following the threat model of \citet{saha2020hidden} and \citet{souri2022sleeper} as described in \Cref{sec:threat_model}. Details regarding the experimental setup can be found in \Cref{app:exp_setup}. Further experimental evaluations are provided in \Cref{app:additional_exp}.

\begin{table*}[t]
\caption{\textbf{Targeted data poisoning. GDP achieves a far higher success rate than Witches' Brew, even with only a small budget}. Experiments are conducted on ImageNet with ResNet-18 models. We see that injection of only 50-100 poisoned samples is enough for the attack to be effective.} 
\label{tab:baseline_imagenet_poisoning}

\centering 
\resizebox{\textwidth}{!}{
    \begin{tabular}{l@{\hspace{7mm}}c@{\hspace{7mm}}c@{\hspace{7mm}}c} 
    \toprule 
    \textbf{Poison budget} &\textbf{Perturbation $\ell_{\infty}$-norm} &\textbf{Witches' Brew} & \textbf{GDP (ours)} \\
    \midrule
    $50$ images ($\sim 0.004\%$) & $8/255$ & $0\%$ & $\textbf{40\%}$  \\
    $100$ images ($\sim 0.008\%$) & $8/255$ & $10\%$ & $\textbf{50\%}$  \\
    \midrule
    $50$ images ($\sim 0.004\%$) & $16/255$ & $0\%$ & $\textbf{50\%}$  \\
    $100$ images ($\sim 0.008\%$) & $16/255$ & $20\%$ & $\textbf{80\%}$  \\
    \bottomrule 
    \end{tabular}
}

\end{table*}

\begin{table*}[t]
\caption{\textbf{Backdoor attacks. GDP achieves a far higher success rate than existing backdoor attacks, even with only a small budget}. Experiments are conducted on CIFAR-10 with ResNet-18 models. Perturbations are bounded in $\ell_{\infty}$-norm  by $16/255$. We see that injection of only 25-50 poisoned samples is enough for the attack to be effective.} 
\label{tab:baseline_cifar_backdoor}

\centering 
\resizebox{\textwidth}{!}{
    \begin{tabular}{l@{\hspace{10mm}}c@{\hspace{10mm}}c@{\hspace{10mm}}c@{\hspace{10mm}}c} 
    \toprule 
    \textbf{Poison budget} &\textbf{CLBA} & \textbf{HTBA} &\textbf{Sleeper Agent} & \textbf{GDP (ours)} \\
    \midrule
    $25$ images ($0.05\%$) & $0.85\%$ & $1.09\%$ & $13.44\%$ & $\textbf{35.28\%}$  \\
    $50$ images ($0.1\%$) & $0.98\%$ & $1.27\%$ & $29.11\%$ & $\textbf{47.71\%}$  \\
    \bottomrule 
    \end{tabular}
}

\end{table*}

\begin{table*}
\caption{\textbf{Improving existing targeted data poisoning attacks with GDP base samples.} Experiments are conducted on CIFAR-10 with 6-layer ConvNets, and perturbations are bounded in $\ell_{\infty}$-norm by $32/255$. GDP base samples are generated using the corresponding downstream poisoning loss functions.} 
\label{tab:poisoning_GDP}

\centering 
\resizebox{\textwidth}{!}{
    \begin{tabular}{l@{\hspace{5mm}}c@{\hspace{5mm}}c@{\hspace{5mm}}c@{\hspace{5mm}}c} 
    \toprule 
    \textbf{Poison budget} &\textbf{Poison Frogs} & \textbf{Bullseye} &\textbf{Poison Frogs $+$ GDP} & \textbf{Bullseye $+$ GDP} \\
    \midrule
    $100$ images ($0.2\%$) & $10\%$ & $20\%$ & $\textbf{20\%}$ & $\textbf{40\%}$  \\
    $200$ images ($0.4\%$) & $10\%$ & $20\%$ & $\textbf{30\%}$ & $\textbf{60\%}$  \\
    \bottomrule 
    \end{tabular}
}

\end{table*}

\begin{table*}[b!]
\caption{\textbf{Improving existing backdoor attacks with GDP base samples.} Experiments are conducted on CIFAR-10 with ResNet-18 models, and perturbations are bounded in $\ell_{\infty}$-norm by $16/255$. GDP base samples are generated using the Sleeper Agent loss function and the victim is fine-tuned on the poisoned training set following \citet{saha2020hidden}.} 
\label{tab:backdoor_GDP}

\centering 
\resizebox{0.6\textwidth}{!}{
    \begin{tabular}{l@{\hspace{10mm}}c@{\hspace{10mm}}c} 
    \toprule 
    \textbf{Poison budget} &\textbf{HTBA} &\textbf{HTBA $+$ GDP} \\
    \midrule
    $50$ images ($0.1\%$) & $1.54\%$ &  $\textbf{17.58\%}$  \\
    \bottomrule 
    \end{tabular}
}

\end{table*}

\begin{table*}
\caption{\textbf{Small perturbations. GDP enables stronger targeted data poisoning under small $\ell_{\infty}$-norm bound perturbations.} Experiments are conducted on CIFAR-10 with ResNet-18 models and poison budget of 50 images (0.1\%).} 
\label{tab:epsilon_cifar_poison}

\centering 
\resizebox{0.75\textwidth}{!}{
    \begin{tabular}{c@{\hspace{10mm}}c@{\hspace{10mm}}c} 
    \toprule 
    \textbf{Perturbation $\ell_{\infty}$-norm} & \textbf{Witches' Brew} & \textbf{GDP (ours)}\\
    \midrule
    $8/255$  & $0\%$ & $\textbf{10\%}$  \\
    $12/255$  & $0\%$ & $\textbf{40\%}$  \\
    \bottomrule 
    \end{tabular}
}

\end{table*}

\begin{table*}
\caption{\textbf{Small perturbations. GDP enables stronger backdoor attacks under small $\ell_{\infty}$-norm bound perturbations.} Experiments are conducted on CIFAR-10 with ResNet-18 models and poison budget of 50 images (0.1\%).} 
\label{tab:epsilon_cifar_backdoor}

\centering 
\resizebox{0.75\textwidth}{!}{
    \begin{tabular}{c@{\hspace{10mm}}c@{\hspace{10mm}}c} 
    \toprule 
    \textbf{Perturbation $\ell_{\infty}$-norm} & \textbf{Sleeper Agent} & \textbf{GDP (ours)}\\
    \midrule
    $8/255$  & $6.91\%$ & $\bf{12.89\%}$  \\
    $12/255$  & $13.58\%$ & $\bf{25.58\%}$  \\
    \bottomrule 
    \end{tabular}
}

\end{table*}

\subsection{Potent Poisons, Even in Small Quantities}
\label{sec:potent_poisons}
As baselines, we compare to the strongest existing targeted data poisoning attack, Witches' Brew \citep{geiping2020witches}, along with Poison Frogs \citep{shafahi2018poison} and Bullseye Polytope \citep{aghakhani2021bullseye} in \Cref{tab:baseline_cifar_poisoning}. These attacks usually require several hundred poisoned samples to be effective on CIFAR-10; providing only a budget of 25 and 50 poisoned images was previously insufficient to attack the model. However, we find that potent poisons developed with our approach described in \Cref{sec:method} are far more effective, even though 25 images constitute only $0.05\%$ of the CIFAR-10 training set.

We further find that these poisoning results scale to large-scale experiments on ImageNet, as shown in \Cref{tab:baseline_imagenet_poisoning}. Again modifying only a tiny subset of training images ($0.004\%$-$0.008\%$) is sufficient to poison the model, exceeding the effectiveness of existing approaches by a very wide margin.

This success also extends to backdoor attacks, where we show results in \Cref{tab:baseline_cifar_backdoor}, noting the effectiveness of the attack compared to a recent state-of-the-art hidden trigger backdoor attack, Sleeper Agent \citep{souri2022sleeper}, as well as Clean-Label Backdoor Attacks (CLBA) \citep{turner2018clean} and Hidden-Trigger Backdoor Attacks (HTBA) \citep{saha2020hidden}. GDP outperforms all of the above backdoor attacks. Backdoor attack evaluations on ImageNet can be found in \Cref{app:imagenet}.


We can also use our diffusion-generated images as base samples for other existing targeted poisoning and backdoor attacks for significant boosts in effectiveness. In \Cref{tab:poisoning_GDP} and \Cref{tab:backdoor_GDP}, we see that GDP base samples massively improve the success rates of Poisons Frogs and Bullseye Polytope targeted data poisoning attacks as well as Hidden-Trigger Backdoor Attacks.

\begin{table*}
\caption{\textbf{Black-box targeted poisoning attacks. GDP improves targeted poisoning success rates in the harder black-box setting.} Perturbations have $\ell_{\infty}$-norm bounded above by $16/255$.  Poisons crafted using ResNet-18 surrogate and transferred to different victim architectures.  Experiments are conducted on CIFAR-10.} 
\label{tab:transfer_poisoning}
\centering 
\resizebox{\columnwidth}{!}{
\begin{tabular}{l@{\hspace{10mm}}c@{\hspace{10mm}}c@{\hspace{10mm}}c} 
\toprule 
\textbf{Victim architecture} &\textbf{Poison budget} &\textbf{Witches’ Brew} & \textbf{GDP (ours)} \\

\midrule
ResNet-34 & $25$ images ($0.05\%$) & $2.50\%$ & $\textbf{30.00\%}$ \\
ResNet-34 & $50$ images ($0.1\%$) & $20.00\%$ & $\textbf{37.50}\%$ \\
ResNet-34 & $100$ images ($0.2\%$) & $35.00\%$ & $\textbf{52.50\%}$ \\
\midrule
MobileNet-V2 & $100$ images ($0.2\%$) & $0\%$ & $\textbf{13.75\%}$ \\

\bottomrule 
\end{tabular}
}
\end{table*}

\begin{table*}[b!]
\caption{\textbf{Black-box backdoor attacks. GDP improves backdoor attack success rates in the harder black-box setting.} Perturbations have $\ell_{\infty}$-norm bounded above by $16/255$.  Poisons crafted using ResNet-18 surrogate and transferred to different victim architectures.  Experiments are conducted on CIFAR-10.} 
\label{tab:transfer_backdoor}
\centering 
\resizebox{\columnwidth}{!}{
\begin{tabular}{l@{\hspace{5mm}}c@{\hspace{5mm}}c@{\hspace{5mm}}c@{\hspace{5mm}}c@{\hspace{5mm}}c} 
\toprule 
\textbf{Attack} &\textbf{Poison budget} &\textbf{MobileNet-V2} & \textbf{VGG11} & \textbf{ResNet-34} & \textbf{Average} \\
\midrule
Sleeper Agent & $25$ images ($0.05\%$) & $5.52\%$ & $7.71\%$ & $11.32\%$ & $8.18\%$   \\
GDP (ours) & $25$ images ($0.05\%$) & $\bf{10.74\%}$ & $\bf{9.77\%}$ & $\bf{26.16\%}$ & $\bf{15.58\%}$   \\
\midrule
Sleeper Agent & $50$ images ($0.1\%$) & $9.20\%$ & $11.20\%$ & $15.74\%$ & $12.38\%$   \\
GDP (ours) & $50$ images ($0.1\%$) & $\bf{13.24\%}$ & $\bf{12.62\%}$ & $\bf{39.64\%}$ & $\bf{21.38\%}$   \\

\bottomrule 
\end{tabular}
}
\end{table*}

\subsection{Not Only Potent, but Also Stealthy}

In contrast to previous work, GDP attacks are also successful with smaller $\ell_\infty$ constraints, as we highlight in \Cref{tab:epsilon_cifar_poison,tab:baseline_imagenet_poisoning} for targeted data poisoning and \Cref{tab:epsilon_cifar_backdoor} for backdoor attacks. Small perturbation attacks are a crucial application scenario, as higher perturbation budgets, such as $16/255$ are more easily detectable during data inspection, e.g. when the data points are labeled by an annotator. 


\subsection{Not Only Potent, but Also Transferable}
\label{sec:experiments_transfer}

We now investigate how transferable these poisons are in the hard black-box setting, where the architecture of the victim model is unknown to the attacker when crafting poisons. 
\Cref{tab:transfer_poisoning,tab:transfer_backdoor} highlight the transferability of GDP in targeted poisoning and backdoor settings, respectively. Compared to Witches' Brew and Sleeper Agent, our approach boosts transferability. In \Cref{app:ensemble}, we present additional experiments with an ensemble of surrogate models.


\subsection{Defenses and Mitigation Strategies}
\label{sec:experiments_defense}

During a data poisoning or backdoor attack scenario, the victim may deploy a defense mechanism to either filter out suspected poisons or modify the training routine to alleviate the impact of poisoning. We therefore test our potent poisons against several widely adopted existing defense methods.

Spectral Signatures \citep{tran2018spectral} is a representative defense approach that computes the top right singular vector of the covariance matrix of the representation vectors (features) and uses these vectors to compute an outlier score for each input.  Inputs that have scores exceeding the outlier threshold are eliminated from the training set. Differentially private training (DP-SGD) \citep{abadi2016deep,hong2020effectiveness} provides protection against data poisoning by adding calibrated noise to the gradients during the training. This defense ensures that the influence of individual training samples is reduced. However, as demonstrated in \citet{geiping2020witches} and \citet{souri2022sleeper}, this defense mechanism encounters a substantial trade-off between reducing poisoning accuracy and preserving validation accuracy. Recent works suggest that strong data augmentations can be implemented during training to mitigate the poisoning success \citep{borgnia2021dp, schwarzschild2020just}. We evaluate our poisoning and backdoor attacks against mixup \citep{zhang2017mixup}, one of the most effective data augmentation techniques for countering data poisoning attacks \citep{borgnia2021dp}. 

Additionally, we test our GDP backdoor attack against STRIP  \citep{gao2019strip}, Neural Cleanse \citep{wang2019neural}, Adversarial Neuron Pruning (ANP) \citep{wu2021adversarial}, and Anti-Backdoor Learning (ABL) \citep{li2021anti}. STRIP detects incoming backdoor-triggered inputs during testing by deliberately perturbing them and analyzing the entropy of the predicted class distribution. A low entropy suggests the presence of a backdoor input, leading to rejection. Neural Cleanse approximates the backdoor trigger using adversarial perturbations. We leverage this defense mechanism to identify the backdoored class in our attacks through outlier detection. ANP proposes to defend against backdoors by pruning the sensitive neurons through adversarial perturbations to model weights. ABL is another method for mitigating backdoor attacks that identifies suspected training samples with the smallest losses. It then proceeds to unlearn these identified poisoned samples to mitigate the backdoor attack.

Our experimental results, presented in \Cref{tab:defense_poison,tab:defense_backdoor}, indicate that our poisoning and backdoor attacks breach these defenses, consistently maintaining high poisoning accuracy.  It must be noted that, as demonstrated in \citet{geiping2020witches} and \citet{souri2022sleeper}, it is possible for these defensive techniques to highly degrade the poisoning accuracy but simultaneously imposing a significant decrease in validation accuracy. Therefore, in our defense experiments, we adjust the corresponding parameters to ensure a sufficiently high validation accuracy.

\begin{table*}[t!]
\caption{\textbf{Defenses against targeted data poisoning} on CIFAR-10 with ResNet-18 models. Perturbations have $\ell_{\infty}$-norm bounded above by $16/255$ and the attacker can poison $0.1\%$ of training images ($50$ images).} 
\label{tab:defense_poison}

\centering 
\resizebox{\textwidth}{!}{
    \begin{tabular}{l@{\hspace{5mm}}c@{\hspace{5mm}}c@{\hspace{5mm}}c@{\hspace{5mm}}c} 
    \toprule 
    & \textbf{Undefended} &\textbf{Spectral Signatures} &\textbf{DP-SGD} & \textbf{Data Augmentation} \\
    \midrule
    Avg. Poison Success & $70\%$ & $50\%$ & $50\%$ & $40\%$  \\
    Avg. Validation Acc. & $92.16\%$ & $90.24\%$ & $91.34\%$ & $91.39\%$  \\
    \bottomrule 
    \end{tabular}
}

\end{table*}

\begin{table*}[t!]
\caption{\textbf{Defenses against backdoor attacks} on CIFAR-10 with ResNet-18 models. Perturbations have $\ell_{\infty}$-norm bounded above by $16/255$ and the attacker can poison $0.1\%$ of training images ($50$ images).} 
\label{tab:defense_backdoor}

\centering 
\resizebox{\textwidth}{!}{
    \begin{tabular}{l@{\hspace{5mm}}c@{\hspace{5mm}}c@{\hspace{5mm}}c@{\hspace{5mm}}c@{\hspace{5mm}}c@{\hspace{5mm}}c@{\hspace{5mm}}c@{\hspace{5mm}}c} 
    \toprule 
    & \textbf{Undefended} &\textbf{Spectral Signatures} &\textbf{DP-SGD} & \textbf{Data Augmentation} & \textbf{STRIP} & \textbf{NeuralCleanse} & \textbf{ANP} & \textbf{ABL} \\
    \midrule
    Attack Success Rate & $47.71\%$ & $16.21\%$ & $27.65\%$ & $24.40\%$ & $39.21\%$ & $47.71\%$ & $29.89\%$ & $27.56\%$  \\
    Avg. Validation Acc. & $92.15\%$ & $90.13\%$ & $91.47\%$ & $91.66\%$ & $91.96\%$ & $92.15\%$ & $82.52\%$ & $89.95\%$ \\
    \bottomrule 
    \end{tabular}
}

\end{table*}

\subsection{Human Evaluation: GDP Poisons Are Clean-Label}
\label{sec:human_eval}

To ensure that the diffusion-generated base samples are indeed clean-label, we also conduct a human evaluation. We assemble a dataset of 500 GDP base samples crafted with the Witches’ Brew
 and Sleeper Agent objectives and 500 randomly sampled natural images from the CIFAR-10 dataset, encompassing 50 samples for each class in both sets. We ask annotators to classify these images into the $10$ CIFAR-10 classes, and annotators are not told which images are synthetic and which are natural. In \Cref{tab:human_evaluation}, we present the accuracies observed on these sets across annotators. Our analysis demonstrates that their accuracy on GDP base samples is comparable to or even exceeds their accuracy on the natural images, indicating that the synthesized base samples maintain their assigned label as desired.

\begin{table*}[t!] 
\caption{\textbf{GDP base samples are clean-label according to human evaluators.} Human evaluation, accuracy at performing CIFAR-10 10-class classification on real and GDP base samples crafted with the Witches' Brew and Sleeper Agent objectives. Humans perform at least as well classifying GDP base samples as they do on clean CIFAR-10 training samples.} 

\label{tab:human_evaluation}
\centering 
\resizebox{0.7\textwidth}{!}{
\begin{tabular}{l@{\hspace{10mm}}c@{\hspace{10mm}}c} 
\toprule 
& \textbf{Natural images} & \textbf{GDP base samples} \\
\midrule
Classification accuracy & $93\%$ & $96.8\%$ \\
\bottomrule 
\end{tabular}
}
\end{table*}



\section{Limitations and Future Work}
\label{sec:limitations}

\begin{itemize}
\item Our method requires a diffusion model trained specifically on the particular data distribution which is computationally expensive, as is guided diffusion generation.  Can we craft equally effective base samples on a tight budget, perhaps without diffusion models at all?
\item We require the entire training set, or at least a large subset, for training the diffusion model.  Can we instead use a general purpose text-to-image diffusion model with appropriate prompts to avoid training a dataset-specific diffusion model?
\item We generate significantly more poisons than we deploy and then filter them, but generating poisons is expensive, so this procedure is inefficient.  Can we devise a more reliable optimization strategy to avoid filtering?
\end{itemize}
\section{Conclusion}
\label{sec:conclusion}

In this work, we showed that the base samples used for poisoning have a very strong impact on the effectiveness of the resulting poisons.  With this principle in mind, we synthesize base samples from scratch specifically so that they lie near potent poisons.  Our guided diffusion approach amplifies the effects of state-of-the-art targeted data poisoning and backdoor attacks across multiple datasets.




\section*{Acknowledgements}
This work was supported by an ONR MURI grant N00014-20-1-2787, and DARPA GARD. Commercial support was provided by Capital One Bank, the Amazon Research Award program, and Open Philanthropy. Further support was provided by the National Science Foundation (IIS-2212182), and by the NSF TRAILS Institute (2229885).

\medskip

{
\bibliography{main}
\bibliographystyle{plainnat}
}

\clearpage


\appendix

\section{Experimental Details}
\label{app:exp_setup}

\subsection{Experimental Setup}
In this section, we provide the details of our experimental evaluations.  In all poisoning experiments discussed in this paper, we consider \textit{targeted data poisoning attacks} which are designed to cause a particular target test image to be misclassified with an intended label \citep{shafahi2018poison}. In our poisoning attacks, the \textit{Poison Success} rate denotes the percentage of instances where the victim network assigns the \textit{target image} the intended label. Consistent with the methodology outlined by \citet{geiping2020witches}, we choose the poison class to be the same as the intended class. Our reported poison success is based on the average performance across 10 randomly selected target-poison pairs.

For backdoor attacks, the \textit{Attack Success Rate} is the rate at which the victim model misclassifies test images from the \textit{target} class, which have been manipulated to include the trigger, with the intended label. It is noteworthy that, in previous backdoor attack studies, this scenario is often referred to as source-target pairs, indicating that images from the source class are misclassified as the target class \citep{saha2020hidden, souri2022sleeper}. However, to prevent confusion with the terminology used in the targeted data poisoning task (i.e., target-intended), we abstain from employing the source-target analogy in our paper. Furthermore, in our backdoor attack experiments, following the approach outlined by \citet{souri2022sleeper}, we exclusively select poisons from the intended class. Consequently, we consistently use the term ``\textit{target-poison}'' for all our data poisoning and backdoor attack experiments. We report the average attack success rate across 10 trials, with randomly selected target-poison pairs.


\begin{figure*}[b!]
    \centering
    \includegraphics[width=\linewidth]{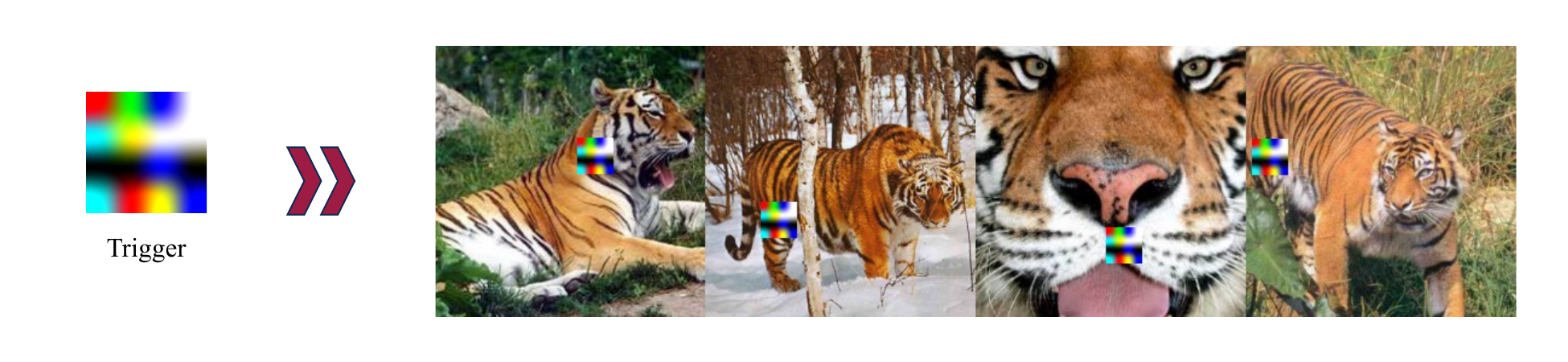}
    \caption{Visualizations of the triggered test images from the ImageNet dataset.}
    \label{fig:trigger}
\end{figure*}

\subsection{Implementation Details}

In the backdoor attack experiments, adhering to the experimental setup outlined by \citet{saha2020hidden} and \citet{souri2022sleeper}, the trigger is a random patch as illustrated in \Cref{fig:trigger}. The patch size is set to $8\times8$ for CIFAR-10 experiments and $30\times30$ for ImageNet experiments. In our baseline experiments, we employ ResNet-18 \citep{he2016deep}, while transfer experiments involve ResNet-34, MobileNet-v2, and VGG11 networks \citep{he2016deep, sandler2018mobilenetv2, simonyan2014very}. Additionally, we consider a 6-layer ConvNet in our  Bullseye and Poison Frogs poisoning attacks following \citet{geiping2020witches}. Our 6-layer ConvNet consists of 5 convolutional layers succeeded by a linear layer. The initial learning rate is set to 0.1 for ResNet-18 and ResNet-34, and 0.01 for ConvNet, MobileNet-v2, and VGG11. All models undergo training for 40 epochs, with the learning rate reduced by a factor of 0.1 at epochs 14, 24, and 35. In line with the approach of \citet{souri2022sleeper}, backdoor attack experiments involve training the victim model for 80 epochs during validation. For all models, we employ SGD with Nesterov momentum and a momentum coefficient of 0.9. Additionally, data augmentation is applied to enhance classifier accuracy, including horizontal flipping with a probability of 0.5 and random crops of size $32\times32$ with zero-padding of 4. For ImageNet, images are resized to $256 \times 256$, followed by a central crop of size $224 \times 224$, horizontal flip with a probability of 0.5, and random crops of size $224 \times 224$ with zero-padding of 28. In all experiments, the victim model is trained from scratch during validation.


To generate GDP base samples, we employ pretrained diffusion models from \citet{DDPM_Ho2020} for CIFAR-10. For ImageNet, we utilize classifier-guided diffusion checkpoints from \citet{dhariwal2021diffusion}, with the classifier scale set to 1. Specifically, for CIFAR-10, we guide the diffusion model to generate samples from the poison class by employing cross-entropy loss calculated using a pretrained ResNet-18 classifier. 

For all experiments, including generating GDP base samples, downstream poisoning, and backdoor attacks, we use one NVIDIA RTX A5000 GPU.

\begin{figure*}[t]
    \centering
    \includegraphics[width=0.85\linewidth]{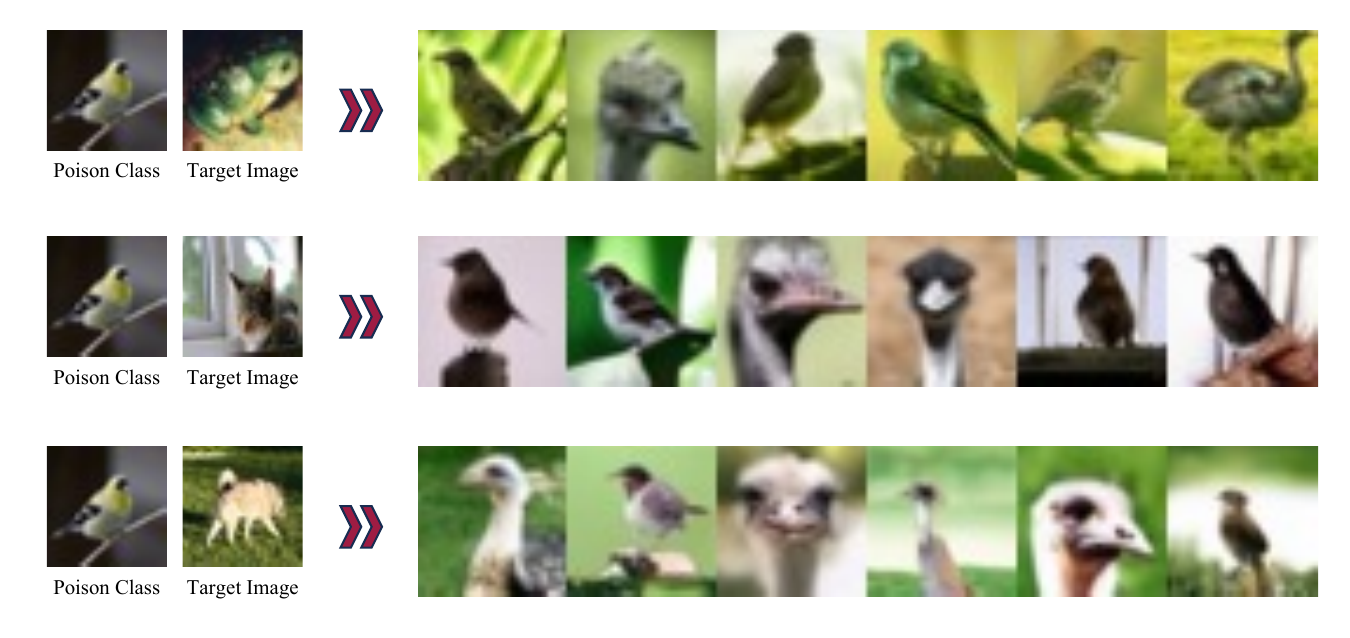}
    \caption{\textbf{The target image visibly influences the corresponding base samples.} We generate base samples from the fixed bird class but using different target images.  We see that the resulting GDP base samples look similar to the target image but remain birds.  Experiments conducted on the CIFAR-10 dataset using the Witches' Brew poisoning objective along with a ResNet-18 model.}
    \label{fig:supp_cifar_bird}
\end{figure*}

\section{Additional Experiments}
\label{app:additional_exp}

\subsection{Effect of GDP on Validation Accuracy}
To demonstrate the effectiveness of GDP, we present the validation accuracy of both clean and poisoned models in both targeted data poisoning and backdoor attack baseline experiments. As shown in \Cref{tab:supp_validation}, it is evident that GDP data poisoning does not lead to a degradation in the validation accuracy of the poisoned model. Furthermore, the validation accuracy of images from the poison class itself remains unaffected.\\

\begin{table*}[t!]

\caption{\textbf{Validation accuracy of GDP}. Experiments are conducted on CIFAR-10 with ResNet-18 models, and perturbations are bounded in $\ell_{\infty}$-norm by $16/255$. Poison budget is 50 images (0.1\%).} 

\label{tab:supp_validation}
\centering 
\resizebox{\linewidth}{!}{
\begin{tabular}{l@{\hspace{10mm}}c@{\hspace{5mm}}c@{\hspace{5mm}}c} 
\toprule 
 & \textbf{Model} & \textbf{Targeted Data Poisoning} & \textbf{Backdoor Attack}  \\
\midrule

Validation accuracy & Clean & $92.08\%$ & $91.96\%$ \\
Validation accuracy & Poisoned & $92.16\%$ & $92.15\%$ \\
Validation accuracy poison class & Clean & $92.15\%$ & $92.36\%$ \\
Validation accuracy poison class & Poisoned & $92.50\%$ & $92.36\%$ \\

\bottomrule 
\end{tabular}
}
\end{table*}

\subsection{More Evaluations on ImageNet}
\label{app:imagenet}

In addition to the backdoor attack experiments detailed in \Cref{sec:potent_poisons}, we further explore backdoor attacks on the ImageNet dataset, a task acknowledged as challenging in prior literature \citep{saha2020hidden,souri2022sleeper}. As demonstrated in \Cref{tab:supp_backdoor_imagenet}, our GDP backdoor attack proves effective even with a small budget of only 100 poisons, while Sleeper Agent yields a negligible attack success rate. It is noteworthy that 100 poisons represents approximately $0.008\%$ of the ImageNet dataset.

\begin{table*}[t!]
\caption{\textbf{Backdoor attacks. GDP achieves a far higher success rate than Sleeper Agent, even with only a small budget}. Experiments are conducted on ImageNet with ResNet-18 models. Perturbations are bounded in $\ell_{\infty}$-norm  by $16/255$. We see that injection of only 100 poisoned samples is enough for the attack to be effective.}

\label{tab:supp_backdoor_imagenet}

\centering 
    \begin{tabular}{l@{\hspace{10mm}}c@{\hspace{10mm}}c} 
    \toprule 
    \textbf{Poison budget} &\textbf{Sleeper Agent} &\textbf{GDP (ours)} \\
    \midrule
    $100$ images ($\sim 0.008\%$) & $3.50\%$ &  $\textbf{14.00\%}$  \\
    \bottomrule 
    \end{tabular}

\end{table*}

\subsection{Additional Transfer Experiments}

\label{app:ensemble}

In \Cref{sec:experiments_transfer}, we discussed how GDP enhances the transferability of poisons across various architectures. To further illustrate this point, we conduct backdoor attack experiments on CIFAR-10 using an ensemble of six models, comprising two ResNet-18, two MobileNet-V2, and two VGG11 models. As depicted in \Cref{tab:supp_transfer_backdoor}, it is evident that, in comparison to the Sleeper Agent, GDP achieves a higher attack success rate by employing ensembling, achieving an average success rate of $19.37\%$ with the budget of only 50 poisons.


\begin{table*}
\caption{\textbf{Transferring backdoor attacks using ensembles.} Experiments are conducted on CIFAR-10 and perturbations have $\ell_{\infty}$-norm bounded above by $16/255$.  Poisons crafted using an ensemble of 6 models. $S$ denotes the size of the ensemble.} 
\label{tab:supp_transfer_backdoor}
\centering 
\resizebox{\textwidth}{!}{
\begin{tabular}{l@{\hspace{7mm}}c@{\hspace{5mm}}c@{\hspace{5mm}}c@{\hspace{5mm}}c@{\hspace{5mm}}c} 
\toprule 
\textbf{Attack} &\textbf{Poison budget} & \textbf{ResNet-18} &\textbf{MobileNet-V2} & \textbf{VGG11}  & \textbf{Average} \\
\midrule
Sleeper Agent ($S=6$) & $25$ images ($0.05\%$) & $10.20\%$ & $6.35\%$ & $10.38\%$ & $8.97\%$   \\
GDP ($S=6$) & $25$ images ($0.05\%$)  & $\textbf{19.73\%}$ & $\textbf{14.56\%}$ & $\textbf{15.95\%}$ & $\textbf{16.74\%}$   \\
\midrule
Sleeper Agent ($S=6$) & $50$ images ($0.1\%$) & $20.25\%$ & $7.35\%$ & $14.89\%$ & $14.16\%$ \\
GDP ($S=6$) & $50$ images ($0.1\%$) & $\textbf{24.16\%}$ & $\textbf{14.65\%}$ & $\textbf{19.30\%}$ & $\textbf{19.37\%}$     \\

\bottomrule 
\end{tabular}
}
\end{table*}



\section{Visualizations}
\label{app:vis}

\begin{figure*}
    \centering
    \includegraphics[width=\linewidth]{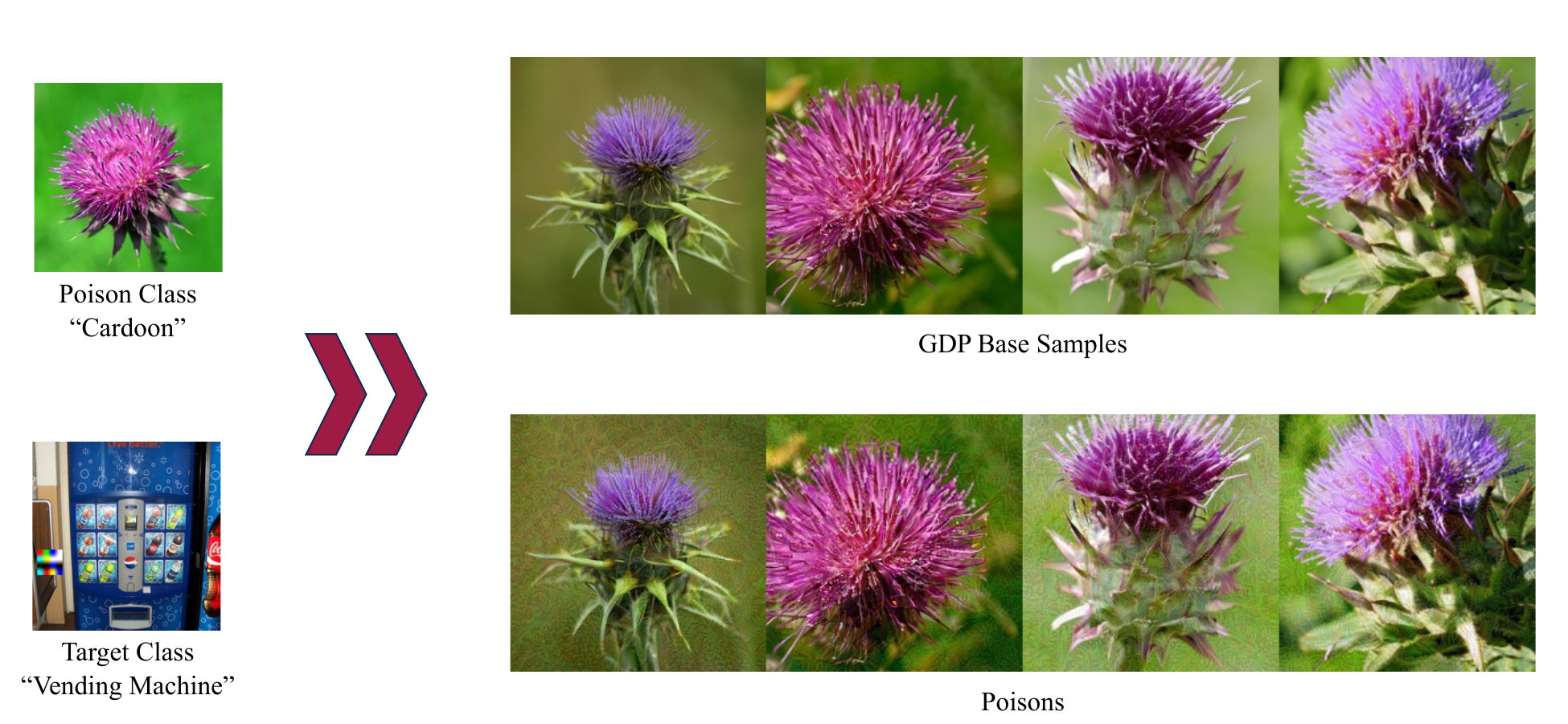}
    \caption{\textbf{GDP base samples and their corresponding poisons (ImageNet).} Experiments conducted using the Sleeper Agent gradient-matching objective with a ResNet-18 model on ImageNet over randomly sampled poison class and target class pairs. Perturbations have $\ell_{\infty}$-norm bounded above by $16/255$, and the patch size is $30\times30$. }
    \label{fig:supp_imageNet_backdoor_1}
\end{figure*}

\begin{figure*}
    \centering
    \includegraphics[width=\linewidth]{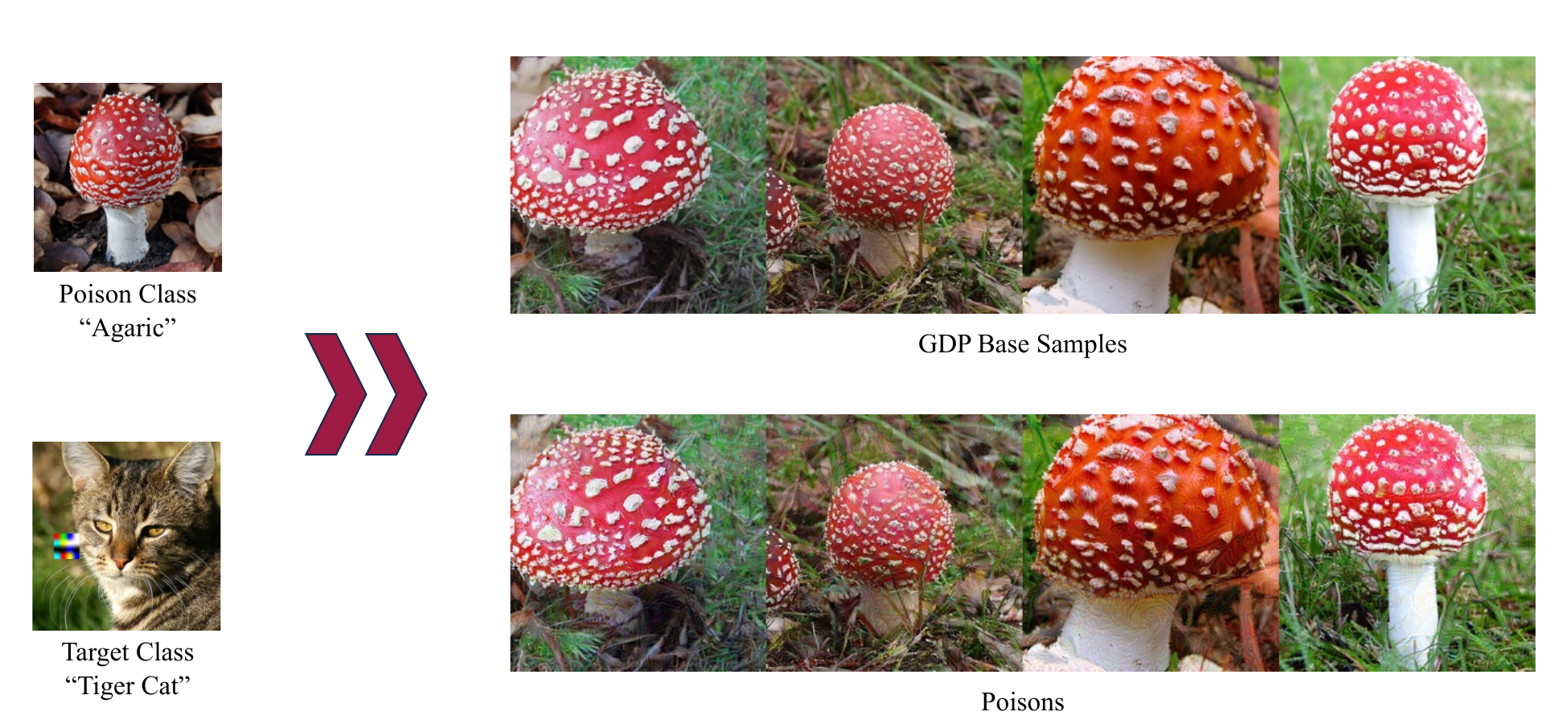}
    \caption{\textbf{GDP base samples and their corresponding poisons (ImageNet).} Experiments conducted using the Sleeper Agent gradient-matching objective with a ResNet-18 model on ImageNet over randomly sampled poison class and target class pairs. Perturbations have $\ell_{\infty}$-norm bounded above by $16/255$, and the patch size is $30\times30$. }
    \label{fig:supp_imageNet_backdoor_2}
\end{figure*}

\begin{figure*}
    \centering
    \includegraphics[width=\linewidth]{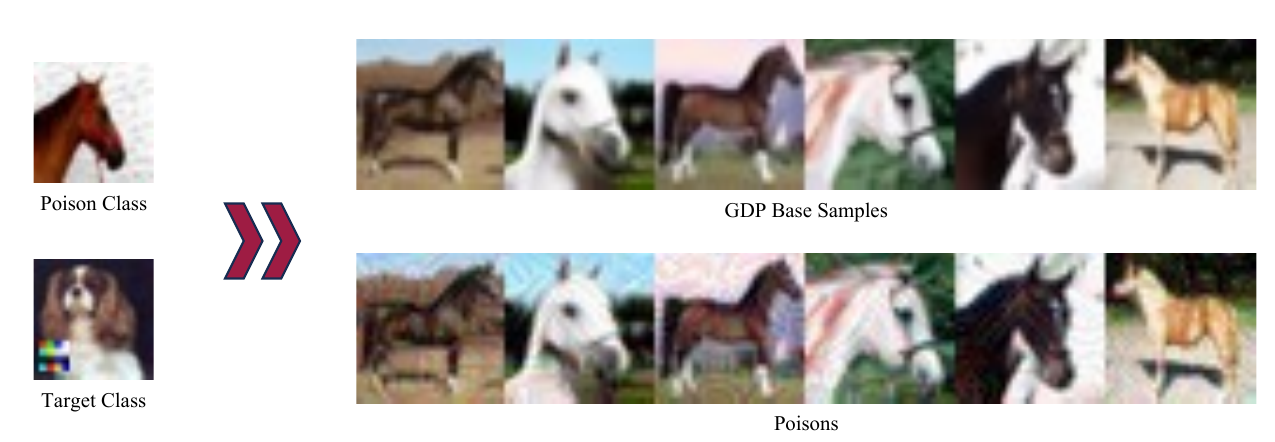}
    \caption{\textbf{GDP base samples and their corresponding poisons (CIFAR-10).} Experiments conducted using the Sleeper Agent gradient-matching objective with a ResNet-18 model on CIFAR-10 over randomly sampled poison class and patched class pairs. Perturbations have $\ell_{\infty}$-norm bounded above by $16/255$, and the patch size is $8\times8$. }
    \label{fig:supp_cifar_vis_3}
\end{figure*}

\begin{figure*}
    \centering
    \includegraphics[width=\linewidth]{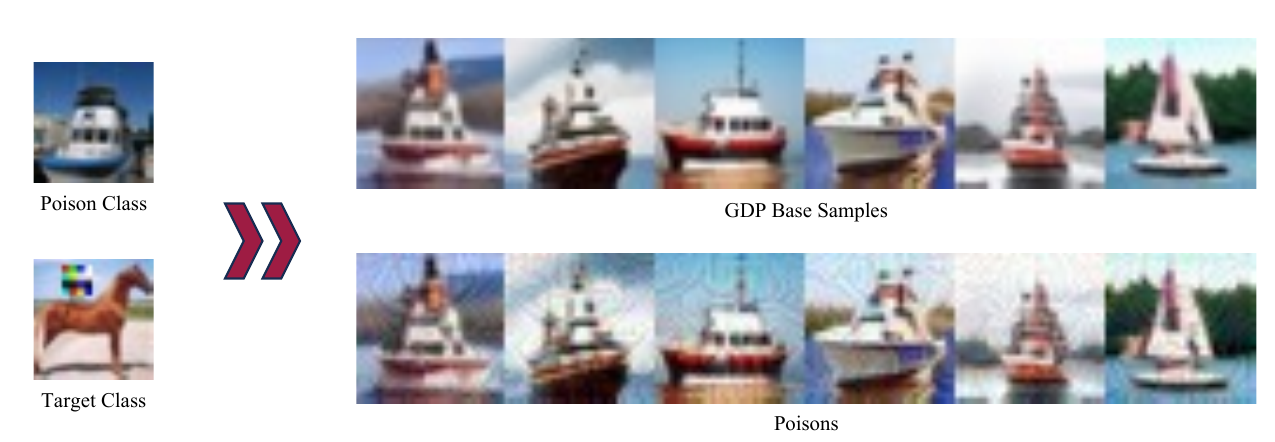}
    \caption{\textbf{GDP base samples and their corresponding poisons (CIFAR-10).} Experiments conducted using the Sleeper Agent gradient-matching objective with a ResNet-18 model on CIFAR-10 over randomly sampled poison class and patched class pairs. Perturbations have $\ell_{\infty}$-norm bounded above by $16/255$, and the patch size is $8\times8$.}
    \label{fig:supp_cifar_vis_4}
\end{figure*}

\begin{figure*}
    \centering
    \includegraphics[width=\linewidth]{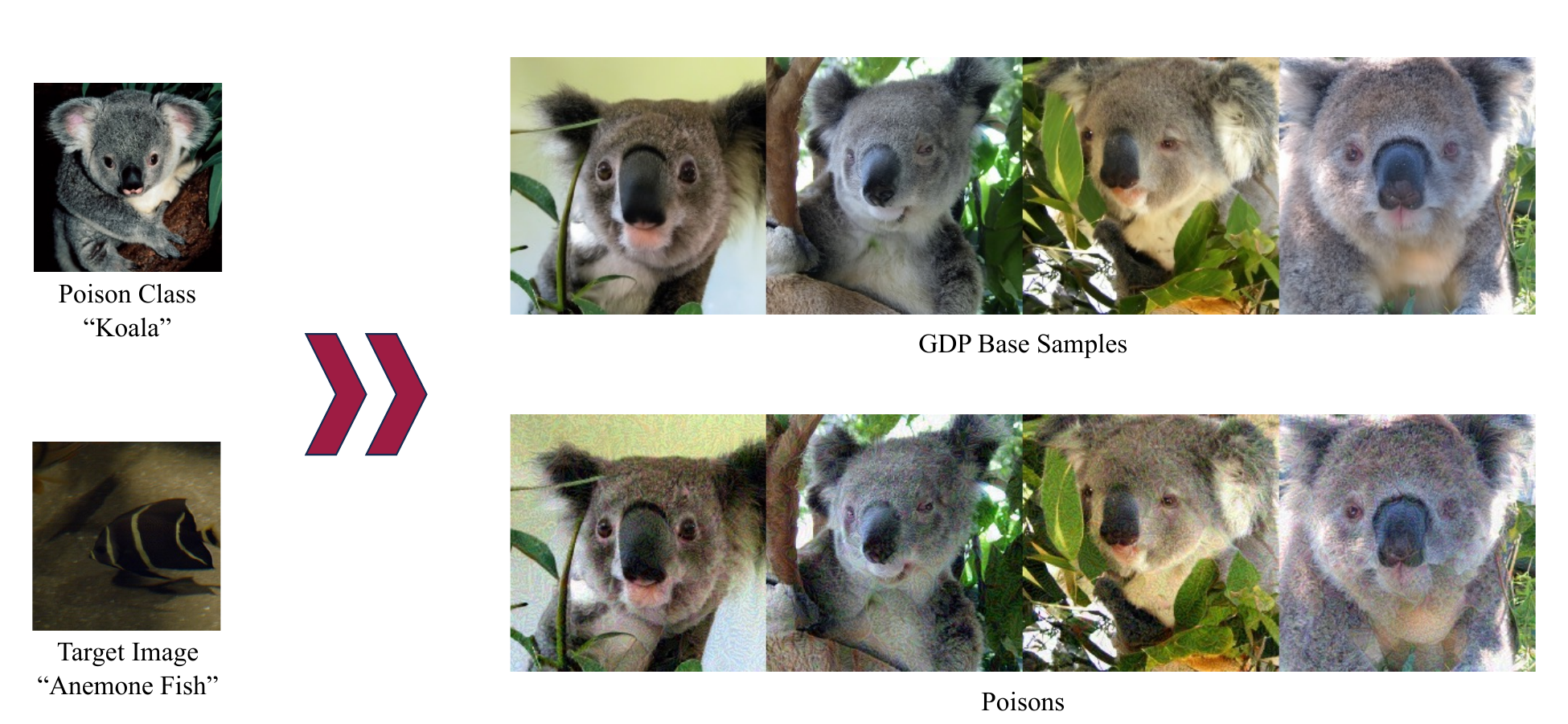}
    \caption{\textbf{GDP base samples and their corresponding poisons (ImageNet).} Experiments conducted using the Witches' Brew gradient-matching objective with a ResNet-18 model on ImageNet over randomly sampled poison class and target image pairs. Perturbations have $\ell_{\infty}$-norm bounded above by $16/255$.}
    \label{fig:supp_imagenet_vis_1}
\end{figure*}

\begin{figure*}
    \centering
    \includegraphics[width=\linewidth]{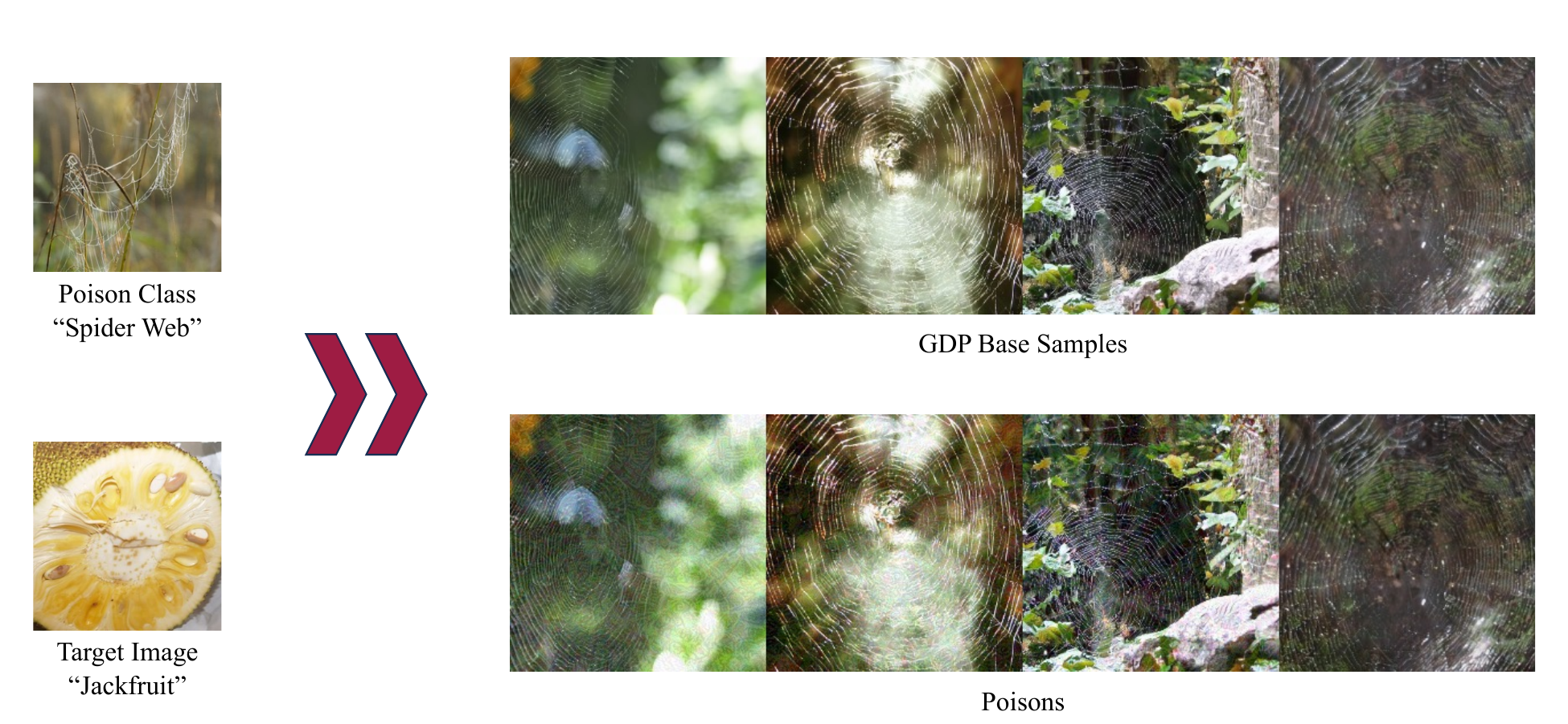}
    \caption{\textbf{GDP base samples and their corresponding poisons (ImageNet).} Experiments conducted using the Witches' Brew gradient-matching objective with a ResNet-18 model on ImageNet over randomly sampled poison class and target image pairs. Perturbations have $\ell_{\infty}$-norm bounded above by $16/255$.}
    \label{fig:supp_imagenet_vis_2}
\end{figure*}

\begin{figure*}
    \centering
    \includegraphics[width=\linewidth]{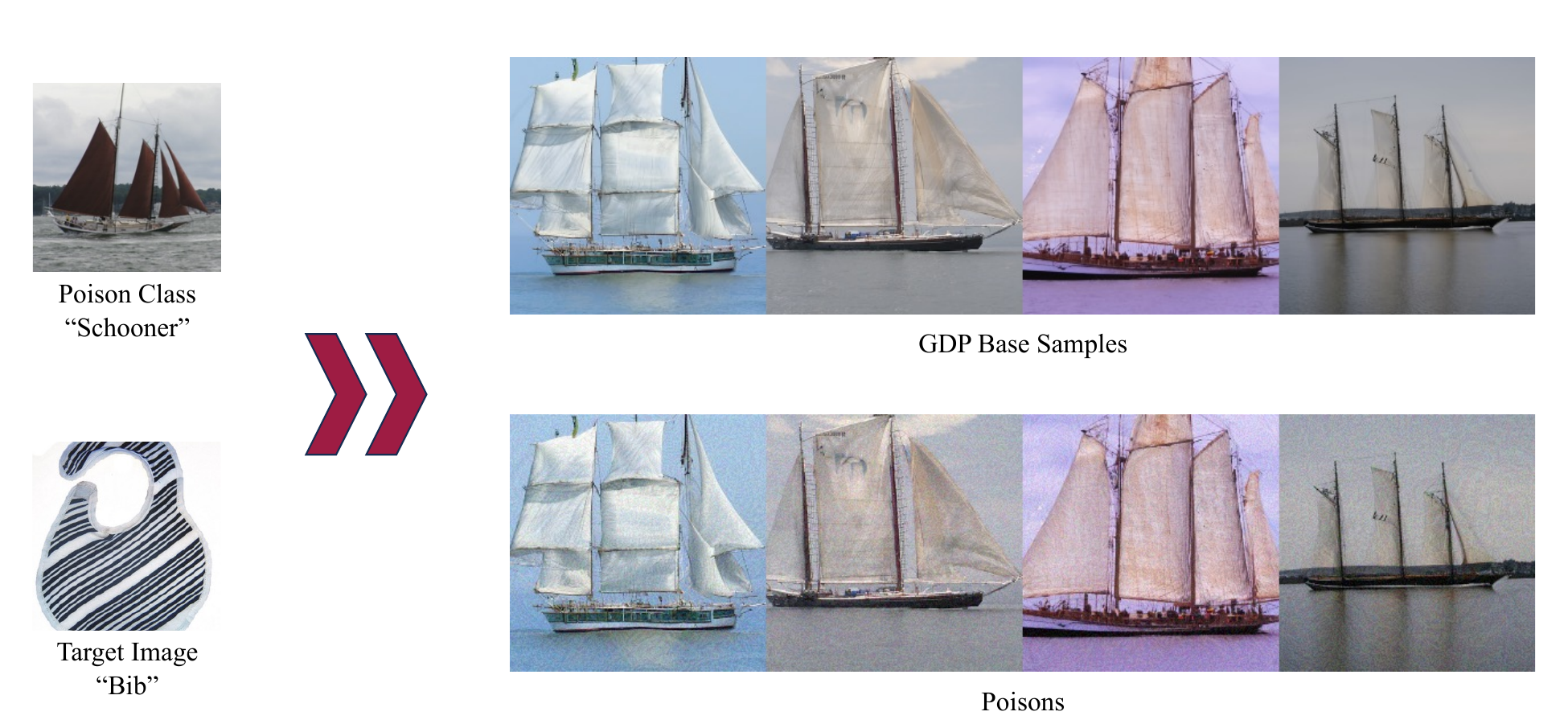}
    \caption{\textbf{GDP base samples and their corresponding poisons (ImageNet).} Experiments conducted using the Witches' Brew gradient-matching objective with a ResNet-18 model on ImageNet over randomly sampled poison class and target image pairs. Perturbations have $\ell_{\infty}$-norm bounded above by $16/255$.}
    \label{fig:supp_imagenet_vis_3}
\end{figure*}

\begin{figure*}
    \centering
    \includegraphics[width=\linewidth]{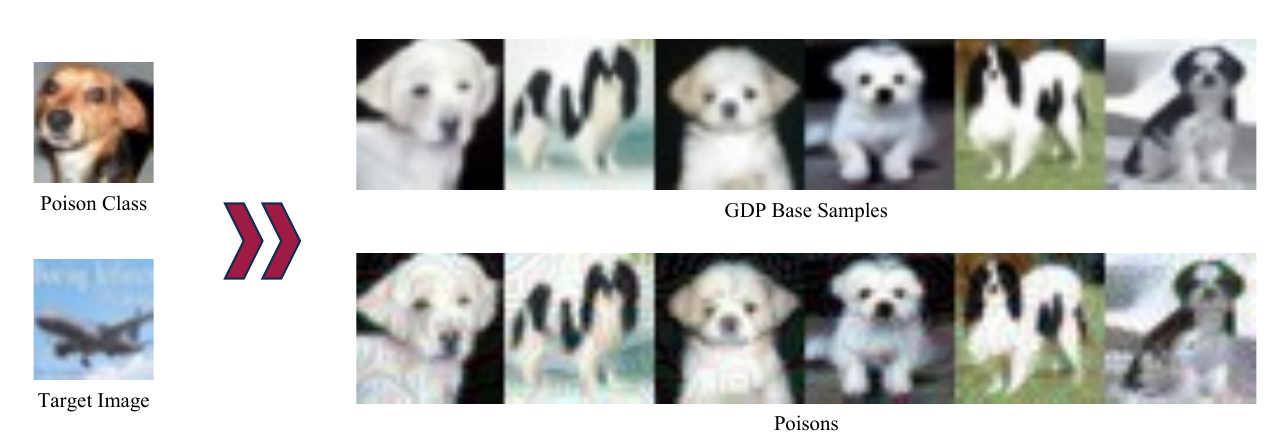}
    \caption{\textbf{GDP base samples and their corresponding poisons (CIFAR-10).} Experiments conducted using the Witches' Brew gradient-matching objective with a ResNet-18 model on CIFAR-10 over randomly sampled poison class and target image pairs. Perturbations have $\ell_{\infty}$-norm bounded above by $16/255$.}
    \label{fig:supp_cifar_vis_1}
\end{figure*}

\begin{figure*}
    \centering
    \includegraphics[width=\linewidth]{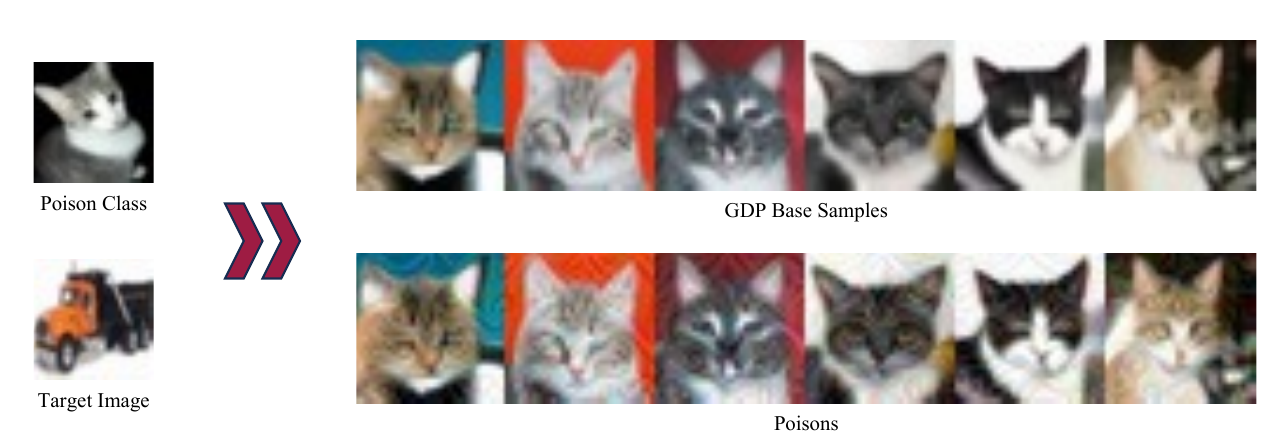}
    \caption{\textbf{GDP base samples and their corresponding poisons (CIFAR-10).} Experiments conducted using the Witches' Brew gradient-matching objective with a ResNet-18 model on CIFAR-10 over randomly sampled poison class and target image pairs. Perturbations have $\ell_{\infty}$-norm bounded above by $16/255$.}
    \label{fig:supp_cifar_vis_2}
\end{figure*}

In this section, we present additional visualizations of GDP attacks on CIFAR-10 and ImageNet datasets. 
In \Cref{fig:supp_cifar_bird}, we observe the influence of a specific target image on the corresponding base samples, even within a fixed poison class. Figures \ref{fig:supp_imageNet_backdoor_1}, \ref{fig:supp_imageNet_backdoor_2}, \ref{fig:supp_cifar_vis_3}, \ref{fig:supp_cifar_vis_4} depict GDP base samples along with their corresponding poisons in backdoor attacks on ImageNet and CIFAR-10. Additionally, GDP base samples and their corresponding poisons in targeted data poisoning attacks on ImageNet and CIFAR-10 are shown in Figures \ref{fig:supp_imagenet_vis_1}, \ref{fig:supp_imagenet_vis_2}, \ref{fig:supp_imagenet_vis_3}, \ref{fig:supp_cifar_vis_1}, \ref{fig:supp_cifar_vis_2}.

\end{document}